\documentclass[10pt,twocolumn,letterpaper]{article}

\usepackage{iccv}
\usepackage{times}
\usepackage{epsfig}
\usepackage{graphicx}
\usepackage{amsmath}
\usepackage{amssymb}

\usepackage{mathtools}
\usepackage{booktabs}
\usepackage{multirow}
\usepackage{enumitem}
\usepackage{color}

\newcommand{\doublecheck}[1]{\textcolor{black}{#1}}
\newcommand{\keypoint}[1]{\vspace{0.05cm}\noindent\textbf{#1}\quad}
\newcommand{\cut}[1]{}
\usepackage{dsfont}
\usepackage{subcaption}
\usepackage[pagebackref=true,breaklinks=true,letterpaper=true,colorlinks,bookmarks=false]{hyperref}

\iccvfinalcopy 

\def\iccvPaperID{3449} 

\ificcvfinal\fi

\begin{document}
 
\title{Joint Visual Semantic Reasoning: Multi-Stage Decoder for Text Recognition}

\author{Ayan Kumar Bhunia\textsuperscript{1} \hspace{.2cm}  Aneeshan Sain\textsuperscript{1,2}\hspace{.2cm} \hspace{.2cm}  Amandeep Kumar\footnotemark[1] \hspace{.2cm} Shuvozit Ghose\thanks{Interned with SketchX} \hspace{.2cm}\\   Pinaki Nath Chowdhury\textsuperscript{1,2}\hspace{.2cm}  Yi-Zhe Song\textsuperscript{1,2} \\
\textsuperscript{1}SketchX, CVSSP, University of Surrey, United Kingdom.\\
\textsuperscript{2}iFlyTek-Surrey Joint Research Centre
on Artificial Intelligence. 
\\{\tt\small \{a.bhunia, a.sain,  p.chowdhury, y.song\}@surrey.ac.uk}     {\tt\small\{shuvozit.ghose,  kumar.amandeep015\}@gmail.com.}}

\maketitle
\ificcvfinal\thispagestyle{empty}\fi

\begin{abstract}
\vspace{-0.2cm}
Although text recognition has significantly evolved over the years, state-of the-art (SOTA) models still struggle in the wild scenarios due to complex backgrounds, varying fonts, uncontrolled illuminations, distortions and other artifacts. This is because such models solely depend on visual information for text recognition, thus lacking semantic reasoning capabilities. In this paper, we argue that semantic information offers a complimentary role in addition to visual only. More specifically, we additionally utilize semantic information by proposing a multi-stage multi-scale attentional decoder that performs joint visual-semantic reasoning. Our novelty lies in the intuition that for text recognition, prediction should be refined in a stage-wise manner. Therefore our key contribution is in designing a stage-wise unrolling attentional decoder where non-differentiability, invoked by \textit{discretely} predicted character labels, needs to be bypassed for end-to-end training. While the first stage predicts using visual features, subsequent stages refine on-top of it using joint visual-semantic information. Additionally, we introduce multi-scale 2D attention along with dense and residual connections between different stages to deal with varying scales of character sizes, for better performance and faster convergence during training. Experimental results show our approach to outperform existing SOTA methods by a considerable margin.

\cut{on both regular as well as irregular text recognition benchmark datasets.}
\cut{This is akin to humans who rely on semantic parsing to arrive at final prediction -- e.g., it would surely be ``nike'' printed on a pair of trainers, even though the image might indicate ``bike''.}  
 
\end{abstract}





\vspace{-0.7cm}
\section{Introduction}\label{sec:intro}

Text recognition has been a popular area of research \cite{baek2019wrong, li2019show, shi2018aster} for decades thanks to its wide range of commercial applications~\cite{long2018scene}, from translation apps in mixed reality, street signs recognition in autonomous driving to assistive technology for the visually impaired~\cite{biten2019scene}, to name a few.
Significant progress in fundamental deep learning components \cite{jaderberg2015spatial, bahdanau2014neural} alongside sequence-to-sequence learning frameworks \cite{shi2018aster, li2019show, qiao2020seed}, have boosted unconstrained word recognition accuracy (WRA) in recent times.
Despite such developments, state-of-the-art text recognition frameworks \cite{bai2018edit, cheng2017focusing, litman2020scatter, wan2020vocabulary, bhunia2021metahtr, bhunia2019handwriting, bhunia2021text} still struggle in wild scenarios \cite{baek2019wrong, long2018scene} due to complex backgrounds, varying fonts, uncontrolled illuminations, distortions and other artifacts. While machines struggle with a combination of these challenges, humans recognise them easily via joint visual-semantic reasoning.  Therefore, the question in focus is -- how to develop a visual-semantic reasoning skill for text recognition? 


\begin{figure}[t]
\begin{center}
\includegraphics[width=0.90\linewidth]{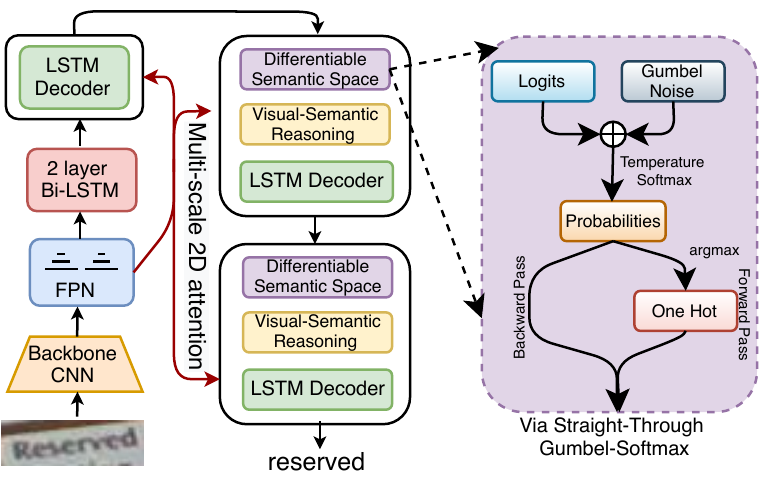} 
\end{center}
\vspace{-0.75cm}
  \caption{{Compared to existing attentional decoder architectures \cite{shi2018aster, li2019show}, we design a novel multi-scale attention decoder for text recognition which is {unpacked in a stage-wise manner} \cut{-- e.g., the first stage unrolls completely across time, only after that second stage starts unrolling and so on}. The problem of non-differentiability, due to discrete-character prediction is bypassed via straight-through Gumbel-Softmax operator \cite{jang2016categorical}, such that the later stages can learn refining strategy over the previous prediction in an end-to-end differentiable way, with joint visual-semantic information.}}
  \vspace{-0.8cm}
\label{fig:Fig1_a.pdf}
\end{figure}

State-of-the-art text recognition systems \cite{baek2019wrong} mostly rely on extracted \textit{visual features} to recognize a word image as a machine readable character-sequence. Follow-up efforts have been made towards improving reasoning ability by increasing the depth of convolutional feature extractor \cite{cheng2017focusing} having larger receptive fields, or introducing pyramidal pooling \cite{wan2020vocabulary} and stacking multiple Bi-LSTM layers \cite{litman2020scatter}. Despite all these attempts that merely lead towards a better context modeling \cite{baek2019wrong}, a semantic reasoning potential \cite{chen2018iterative} is largely missing beyond enriching the visual feature. In wild scenarios, a word image might be blurred, distorted, partly noisy or have artifacts, making recognition extremely difficult using visual feature alone. In such cases, we humans first try to interpret the easily recognizable characters \textit{using visual cues alone}. A semantic reasoning skill is then applied to decode the \textit{final} text by jointly processing the visual and semantic information from previously recognized character sequence. Motivated by this intuition, we propose a novel \emph{multi-stage prediction} paradigm for text recognition. Here the first stage predicts using visual cues, while subsequent stages refine on top of it using joint visual-semantic information, by iteratively \cite{chen2018iterative, CarreiraMalikCVPR2016} building up the estimates.

Designing this joint visual-semantic reasoning framework for text recognition is non-trivial. One might argue that attentional decoder being a sequence-to-sequence model, encapsulates the character dependency \cite{shi2018aster, li2019show, qiao2020seed} and caters for semantic reasoning. However, due to its auto-regressive nature \cite{bahdanau2014neural}, only those characters predicted previously, could provide semantic context at a given step, thus making the semantic context flow unidirectional during inference. While semantic context becomes negligible towards the initial steps, one wrong prediction here would deal a cumulative adverse impact on the later steps (which stays unrefined due to single stage prediction). Therefore, this single stage attentional decoder fails to model the global semantic context, leaving joint visual-semantic reasoning unaccomplished. To explore the entire global semantic context, we need the completely unrolled prediction from first stage, upon which we can build up the global semantic information.  Hence as our first contribution we propose a multi-stage attentional decoder (Figure \ref{fig:Fig1_a.pdf}), where we build up global semantic reasoning on the initial estimate of first stage, which is further refined by subsequent stages.


\doublecheck{Let us consider the word `aeroplane'. For a single stage attentional decoder, if the model predicts `n' instead of `r', `ae\textcolor{red}{n}' would adversely affect rest of the prediction, without any chance of refinement (being single stage). Also, it holds almost negligible semantic context while predicting the first few characters. Considering we unroll the prediction stage-wise, if a character is predicted wrongly, like `ae\textcolor{red}{n}oplane', rest of the characters provide significant context as semantic information. This helps in refining `n' to `r' during the later stages coupled with visual information.}


Moreover, obtaining the prediction from earlier stages, needs a \textit{non-differentiable} \texttt{argmax} operation \cite{jang2016categorical} as characters are discrete tokens. This leads to an inefficient modelling of influence of a prior stage on the next predictions. An apparent approach here might be to adapt teacher forcing \cite{lamb2016professor} for the later stages during training. The later stages intend to learn \emph{how to refine} the initial (might be incorrect) hypothesis towards a correct prediction. This motivation however is defeated on feeding exact ground-truth labels as teacher forcing for subsequent stages. Consequently, we make use of Gumbel-Softmax operation \cite{jang2016categorical} bypassing non-differentiability, and making the network end-to-end trainable even across stages.

In summary our contributions are: 
\textit{First and foremost}, we propose a multi-stage character decoding paradigm with stage-wise unrolling. While the first stage predicts using visual features, subsequent stages refine on-the-top of them using joint visual-semantic information.
\textit{Secondly}, we employ a Gumbel-softmax layer to make visual-to-semantic embedding layer differentiable. The model thus learns its refining strategy from  initial to final prediction in an end-to-end manner. \textit{Thirdly,} from the architectural design,  we introduce multi-scale 2D attention to deal with varying scales of character size, and empirically found adding dense and residual connection between different stages stabilize training for better performance leading to outperforming other state-of-the-arts significantly on benchmark datasets.

\section{Related Works}

\noindent \textbf{Text Recognition:}\cut{ A comprehensive survey on recent deep learning based text recognition methods has been done by Long \etal \cite{long2018scene}. Earlier deep learning based text recognition pipelines were restricted to dictionary words \cite{BissacoICCV2013, JaderbergARXIV2014} and limited by the requirement of costly annotation of character level localization \cite{Jaderberg2015} for training. Later on, these shortcomings were bypassed by connectionist temporal classification (CTC) layer \cite{GravesICML2006} enabling sequence discriminative learning, without any character localized annotation.
Connectionist Temporal Classification (CTC) layer \cite{GravesICML2006} enables sequence discriminative learning, without any character localized annotation \cite{BissacoICCV2013, JaderbergARXIV2014}.} 
While connectionist temporal classification (CTC) layer \cite{GravesICML2006} does not model dependency in the output character space \cite{shi2016rare}, an attention based decoder \cite{shi2018aster} encases language modeling, weakly supervised character detection and character recognition in a single paradigm. Following some seminal works \cite{shi2018aster, lee2016recursive}, attention based decoder became state-of-the-art pipeline for text recognition which includes four successive modules: i) a rectification network \cite{shi2018aster} to simplify irregular text image, ii) convolutional encoder for feature extraction, iii) Bi-LSTM layer for context modeling, and iv) an attentional decoder predicting the characters autoregressively.

Furthermore, the motivation of recent followed-up works can broadly be classified into following directions: (i) \textit{improve rectification network} by introducing iterative \cite{bhunia2021unseen} pipeline \cite{ESIR2019} and modelling geometrical attributes \cite{yang2019symmetry} of text image; (ii) four directional feature encoder \cite{cheng2018aon} for \textit{better convolutional feature extraction}; (iii) \textit{improving attention mechanism} by extending to 2-D attention \cite{li2019show} and hard character localized annotation \cite{cheng2017focusing, liao2019scene}, to better guide the attention based character alignment process. \cut{; even though the later approaches get criticized due to costly annotation requirement of each character's location.} (iv) Recently, stacking multiple Bi-LSTM layers \cite{litman2020scatter} and pyramidal pooling \cite{wan2020vocabulary} on convolutional feature were employed towards the goal of \emph{better context modeling}. These approaches however mainly focus on exploiting visual features, via different architectural modifications  \cite{yanplugnet, yue2020robustscanner} on top of Shi \etal \cite{shi2018aster}, but mostly lack in any semantic reasoning capabilities.

Although some works claim to model semantic reasoning by stacking additional Bi-LSTM layers \cite{litman2020scatter, wan2020vocabulary}, it only helps in modelling better contextual information without having actual reasoning potential. In this context, word-embeddings \cite{qiao2020seed} from pre-trained language model were used to initialize the hidden state of attentional decoder, however we are skeptical towards this. For e.g. two related words ``Chair" and ``Table" may lie close in word-embedding space, but their character combination is way apart, thus questioning usage  of word-embedding for text recognition. Yu \etal's \cite{yu2020towards} architectural design in this direction, gets severely limited on using \texttt{argmax} operation in visual-to-semantic embedding layer which invokes non-differentiability, restricting gradient flow from final prediction layer through this block; making learning deficient (Section \ref{analysis}). \doublecheck{To our belief, ours is the first work employing a fully-differentiable semantic reasoning block that caters multi-stage refining objective for discrete character sequence prediction task.}


\noindent \textbf{Multi-Scale Learning:} This learning paradigm is widely prevalent in object detection \cite{lin2017feature}, recognition \cite{kong2016hypernet, bell2016inside, liu2016ssd} and semantic segmentation \cite{long2015fully, hariharan2015hypercolumns}. Instead of solely relying on low resolution, semantically strong features, multi-scale framework sike MSCNN \cite{cai2016unified}, DAG-CNNs \cite{yang2015multi}, and FPN \cite{lin2017feature}  combine them with high-resolution, semantically weak features for object detection across a diverse range of shape and sizes. We couple multi-scale feature to  generate multi-scale attention vectors for text recognition. 


\noindent \doublecheck{\textbf{Multi-Stage Frameworks:} In spite of computational overhead, multi-stage framework has gained popularity in computer vision task like pose estimation \cite{ ramakrishna2014pose}, object detection \cite{chen2018iterative} and action recognition \cite{farha2019ms} for significantly improved performance. Specifically, Convolutional Pose Machine \cite{wei2016convolutional} is one of the most successful and widely accepted multi-stage deep frameworks for pose-estimation.}


\noindent \textbf{Joint Visual-Semantic Learning:} 
Recently, Graph Convolution Networks~\cite{kipf2016semi} achieved success in object detection \cite{chen2018iterative}, image-text matching \cite{li2019visual}, image captioning \cite{krishna2017visual} by generating enhanced visual features with local and global semantic relationship. In our work, we use 
transformer network \cite{vaswani2017attention} for joint visual 
semantic reasoning.

\begin{figure*}[t]
\begin{center}
\includegraphics[width=\linewidth]{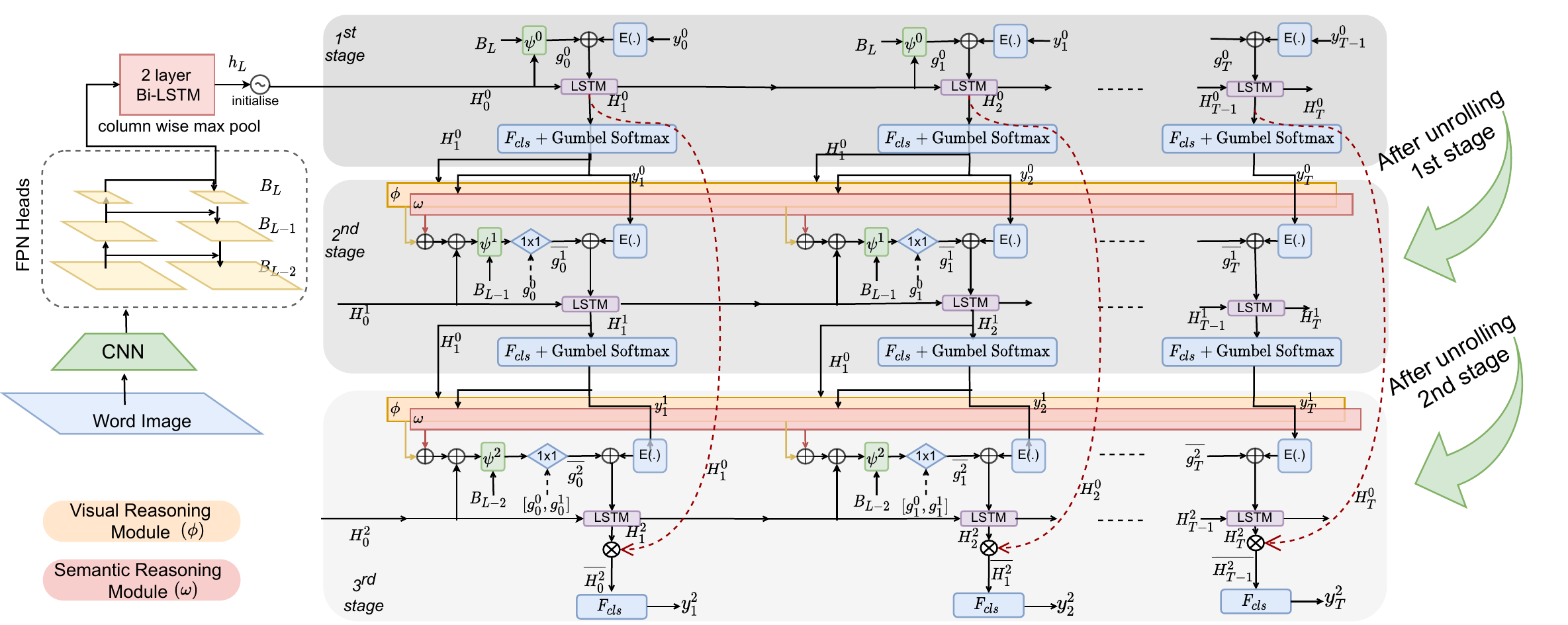} 
\end{center}
\vspace{-.25in}
  \caption
  {
  With the extracted context-rich holistic feature ($h_L$) and multi-scale feature maps ($B_{L}, B_{L-1}, B_{L-2}$), a multi-stage attentional decoder predicts the character sequences in consecutive stages. Once the previous stage's decoder \textbf{completely unrolls itself across time}, the current one begins prediction using the global-semantic information from previous stage's predicted character sequence, coupled with visual features refined via joint visual-semantic reasoning. [$\oplus$ = concatenation ;  $\otimes$ = residual connection with  \textit{LayerNorm}]. Best viewed when zoomed. 
  }
  
\vspace{-.25in}
\label{fig:Fig2}
\end{figure*}

\section{Methodology}

\vspace{-0.1cm}

\keypoint{Overview:} Given an input word image $I$, we intend to predict the character sequence $Y = \{ y_{1},  y_{2}, ...,  y_{T}\}$, where $T$ denotes the variable length of text. Our framework is two-fold: \emph{(i)} a \emph{visual feature extractor} extracts context-rich holistic feature and multi-scale feature maps. \emph{(ii)} Following that, a \emph{multi-stage attentional decoder} builds up the character sequence estimates, in a stage-wise successive manner.
While dealing with irregular/curved word images \cite{yang2019symmetry, cheng2018aon}, image rectification based approaches \cite{yang2019symmetry} often fall short \cite{cheng2018aon, liao2019scene}. To do away with the burden of adding a separate sophisticated rectification network entirely, we follow a 2D attention mechanism \cite{li2019show} that helps to localize individual character in a weakly-supervised manner during decoding.

\subsection{Visual Feature Extraction} \label{basemodel}
We adopt ResNet from \cite{shi2018aster} as a backbone convolutional network to extract visual features from input image. To deal with characters of varying scales, we extend to multi-scale architecture for text recognition, with the help of Feature Pyramid Networks \cite{lin2017feature} which makes every resolution level semantically strong using lateral connections. Let a feature-map from particular scale be represented as  $B_{l} \in \mathbb{R}^{H_{l}\times W_{l}\times D}$; where $l=L$ denotes deepest residual block having lowest resolution but highest level semantics. $H_{l}$ and $W_{l}$ are the height and width of the feature map from respective scales which depend on the accumulated strides of successive pooling layers, with all scales having $D$ channels uniformly \cite{lin2017feature}. To balance between computational ease and performance gain, we consider ${l = \{L, L-1, L-2\}}$ through empirical validation. Visual features have two components,
\emph{(i)} \emph{multi-scale feature-maps} $\mathbf{\{B_{L}, B_{L-1}, B_{L-2}\}}$ which acts as context for 2D attention in the later decoding process. 
\emph{(ii)} The \emph{holistic feature} $\mathbf{h_{L}}$, used to initialize the initial state of \textit{first stage} decoder. This $\mathbf{h_{L}}$ is recognised as the final hidden state of a 2-layer Bi-LSTM which takes in a sequential feature ($W_{L} \times D$), obtained from column-wise max-pooling on feature-map $B_{L}$ from deepest residual block (ensuring height stays unity), followed by reshaping.

\subsection{Joint Visual-Semantic Reasoning Decoder}
\keypoint{Overview:} 
Let the prediction from $s^\mathrm{th}$ stage decoder be denoted as 
$Y^{s} = \{  y_{1}^{s},  y_{2}^{s}, ...,  y_{T}^{s}\}$. Specifically, the first-stage decoder relies only on the extracted feature. Subsequent stages additionally use \emph{global semantic information} that is built on top of the initial estimate, in a stage-wise decoding paradigm.
For completeness, we first describe basic attentional decoder in a generalized fashion (ignoring stage notation). Later on we particularly illustrate the design for \textit{first stage} ($s=0$) vs. \textit{later stages} ($s\geq1$).

\vspace{-0.3cm}
\subsubsection{Attentional Decoder Background}
\vspace{-0.1cm}
Text recognition framework aims to model conditional distribution $\mathrm{P(Y|I)}$, which can be factorized as $\mathrm{P(Y|I) = \prod_{t=0}^{T} P(y_{t} | \mathcal{V}, y_{<t})}$ where each character output $y_{t}$ is modelled via conditional distribution over extracted visual information $\mathcal{V}$, and the history of previously predicted characters $y_{<t}$ till then. The basic attentional decoder \cite{shi2018aster} models this factored conditional distribution using an auto-regressive Recurrent Neural Network (RNN) as follows:
\vspace{-0.25cm} 
\begin{equation}\label{eqn_1}
P(Y|I)  = \prod_{t=0}^{T} P(y_{t} |g_{t}, H_{t-1}, y_{t-1}) 
\vspace{-0.25cm}
\end{equation}

Every time-step prediction $\mathrm{y_{t}}$ is conditioned on three factors: (i) $\mathrm{H_{t-1} \in \mathbb{R}^{d_{rnn}}}$ : the previous hidden state of RNN that captures the history knowledge of previously predicted characters $y_{<t} = \{y_{0},	\cdots,  y_{t-1}\}$. (ii) The apparent \textit{influence} of previously predicted character $y_{t-1}$ upon predicting $\mathrm{y_{t}}$, following character modelling protocol. (iii) The \textit{glimpse vector} $g_{t}$, that learns to encode the visual information by attending a smaller \emph{specific} part of visual feature, which is maximally relevant to predict the character $y_{t}$. 
Technically, $\mathbf{g_{t} = \psi(\mathcal{B}, {Q_{t}})}$, where $\mathcal{B}$ is a spatial feature-map, encoding visual information from previous convolutional network, and $Q_{t}$ acts as a query to locate the attentive regions for predicting $y_{t}$. Mathematically put,
 
 \vspace{-0.6cm}
\begin{align}\label{eqn2}
 \begin{cases}
  J =  \mathrm{tanh}( W_{\mathcal{B}} \circledast  \mathcal{B} + W_{H} \otimes Q_{t})   \\
 \alpha_{i,j}  =  \mathrm{softmax}(W_{attn} \otimes J_{i,j}) \\
  g_{t} =  \sum_{i,j} \alpha_{i,j} \cdot \mathcal{B}_{i,j}  \;  i = [1, ..H],  \; j = [1, ..W]  
 \end{cases}
\vspace{-0.8cm}
\end{align}

Here, ``$\circledast $" and ``$\otimes$" denote convolution and matrix multiplication respectively. $W_{B}$, $W_{H}$, $W_{attn}$ are the learnable weights. Usually, $\mathbf{Q_{t} = H_{t-1}}$ containing history of prediction information is used as a query to locate $y_{t}$. Moreover, query vector enriched in global semantic information (e.g. as in $s\geq1$) could also be used instead, for better performance. While calculating the attention weight $\alpha_{i,j}$ at every spatial position $(i,j)$, we employ a convolution operation with $3\times3$ kernel $W_{\mathcal{B}}$ to consider the neighborhood information in 2D attention mechanism.

The current hidden state $H_{t}$ is updated by:  $\mathbf{H_t =  {f_{rnn}}(H_{t-1}; \; [E(y_{t-1}), \; g_{t}] ) )}$, where $E(.)$ is character embedding layer with embedding dimension $\mathbb{R}^{128}$, and [.] signifies a concatenation operation. Finally, we apply a final linear classification layer having learnable weights  ($W_{c}$, $b_{c}$) and giving logits $l_{t} = \mathrm{\mathbf{F_{cls}}(H_{t}); \; l_{t} \in \mathbb{R}^{|V|}}$ where $|V|$ denotes the character vocabulary size.  The current step character is obtained as: $\mathbf{P(y_{t}) = \mathrm{\texttt{softmax}}(l_{t})}$.

\vspace{-0.3cm}
\subsubsection{Decoder Stage $\mathbf{s=0}$}
\vspace{-0.1cm}
Henceforth, we affix notation for specific decoder stage keeping earlier mathematical notation intact. For the first stage decoder RNN $f_{rnn}^{0}$, the initial hidden state is initialized from holistic visual feature: $\mathrm{\mathbf{H_{0}^0 = \mathrm{tanh}(W_{v} \otimes h_{L} + b_{v})}}$, with $W_{v}$, $b_{v}$ being trainable parameters. 
This enriches $f_{rnn}^{0}$ with holistic visual information, while $\mathrm{\mathbf{g_{t}^{0} = \psi^{0}(B_{L}, H_{t-1}^{0})}}$ augments with localized character specific information. At every $t$-th time step,  we obtain the distribution over the output character space as $\mathrm{\mathbf{P(y_{t}^{0}) = \texttt{softmax}(F_{cls}^{0}(H_{t}^{0}))}}$ and $\mathrm{\mathbf{H_t^{0} =  {f_{rnn}^{0}}(H_{t-1}^{0}; \; [E(y_{t-1}^{0}), \; g_{t}^{0}] )}}$. The decoding process stops once the `end-token' is predicted. Sequences having variable length in batches are handled by zero-padding.

\vspace{-0.3cm}
\subsubsection{Decoder Stage  $\mathbf{s\geq1}$}
\vspace{-0.1cm}
 
Incorrect instances might exist in the prediction of preceding stage, which is why the later stages should work towards refining erroneous predictions while keeping the correct ones intact. While this seems similar to Language Model (LM) based post-processing \cite{RozovskayaCorrection} or Error Correction Network \cite{RozovskayaCorrection}, our proposed stage-wise decoders are all coupled in an end-to-end trainable deep architecture.  Here, gradients can backpropagate across stages during training, thus leading to learning better data driven refining strategy \emph{re-utilising} the visual feature.  The later stage decoders $\mathrm{s\geq1}$ are modelled as follows: 
\vspace{-0.25cm}
\begin{equation}\label{eqn3}
P(Y^{s}|I) = \prod_{t=0}^{T} P(y_{t}^{s} |g_{t}^{s}, H_{t-1}^{s}, y_{t}^{s-1},  \mu_{t}^{s})   
\vspace{-0.25cm} 
\end{equation}

Fundamentally, there are three differences compared to basic attentional decoder (Eqn. \ref{eqn_1}):

(i) $y_{t}^s$ is conditioned on \emph{joint visual-semantic information} $\mathbf{\mu_{t}^{s} = [\vartheta_{t}^{s}, \chi_{t}^{s}] }$, where \textit{visual-part} comes from $\mathbf{\vartheta_{t}^{s} =  \phi_{t}(H^{s-1})}$ and \textit{global semantic part} comes from $\mathbf{\chi_{t}^{s} = \omega_{t}(\mathrm{E}(Y^{s-1}))}$. Here, $\phi(\cdot)$ and $\omega(\cdot)$ are \emph{reasoning modules} working on previous stage's character aligned visual feature $H^{s-1} \in \mathbb{R}^{T \times d_{rnn}}$ and semantic characters $Y^{s-1} \in \mathbb{R}^{T \times |V|}$ feeding through character embedding layer $E(\cdot)$ respectively. $\phi_t(\cdot)$ and $\omega_t(\cdot)$ represent $t$-th time step output for respective module. 
Once the previous stage decoder completely unrolls itself, all characters from $Y^{s-1}=\{y_{1}^{s-1}, y_{2}^{s-1}, \cdots , y_{T}^{s-1} \}$ being \textit{concurrently} present, augments a global semantic information for reasoning. The main motive of later stages is to learn a refinement strategy. As we already obtain \emph{character aligned visual-semantic feature} $H^{s-1}=\{H_{0}^{s-1}, H_{1}^{s-1}, \cdots , H_{T}^{s-1} \}$ from the previous stage, we employ a reasoning module to capture enhanced visual reasoning over all the character aligned visual-semantic features from the previous stage. 

(ii) For $g_{t}^{s}$, we additionally use joint visual semantic information $\mathbf{\mu_{t}^{s}}$ for query; thus $\mathbf{Q_{t}^{s} = [\mu_{t}^{s}, H_{t-1}^{s}]}$, and higher resolution feature-map is used as $\mathcal{B} = B_{L-s}$ (e.g., $B_{L-1}$ for $s=1$) to couple multi-scale feature learning in a multi-stage decoder. Thus glimpse vector is $\mathbf{g_{t}^{s} = \psi^{s}(B_{L-s}, [\mu_{t}^{s}, H_{t-1}^{s}])}$.  

(iii) While $s =0$ acts following baseline attentional decoder (Eqn. \ref{eqn_1}), the role for $s\geq1$ is to learn \emph{refining strategy} over previous predictions. Thus instead of feeding previous time-step prediction $y_{t-1}^{s}$, we feed prediction from previous stage corresponding to the same time-step as $y_{t}^{s-1}$.

\keypoint{Differentiable Semantic Space:}   
Obtaining \emph{discrete character token} from distribution over the character vocabulary $P(y_{t})$ requires non-differentiable \texttt{argmax} operation. As our motivation lies in coupling multi-stage decoder in a end-to-end trainable framework, we employ Gumbel-softmax re-parameterisation trick \cite{jang2016categorical} with Straight-Through (ST) gradient estimator such that gradient can backpropagate across stages. This empowers the model to learn \emph{reasoning based refining strategy} over previous prediction. In Gumbel-softmax, discontinuous \texttt{argmax} operation is replaced by a \emph{differentiable} softmax function. Given the output logits $l_{t}^{s} = F_{cls}^{s}(H_{t}^{s})$ and $l_{t} \in \mathbb{R^{|V|}} $, the output probabilities of choosing $j$-th character token are:

\vspace{-0.2cm}
\begin{equation}\label{diff_Sem}
p_{t, j}^{s} = \frac{\exp(l_{t,j}^{s} + g_{t,j}^{s})/\tau}{\sum_{j=1}^{j=|V|}\exp(l_{t,j}^{s} + g_{t,j}^{s})/\tau}
\vspace{-0.2cm}
\end{equation}

\noindent where,  $g_{t,j}^{s}$ represents Gumbel-noise \cite{jang2016categorical}, and $\tau$ is a temperature parameter. During forward pass, it generates one-hot vector $y_{t}^{s} = \{y_{t,1}^{s}, y_{t, 2}^{s}, \cdots, y_{t,|V|}^{s} \} =$ \texttt{Gumbel-Softmax}$(l_{t}^{s})$ where  ${y}_{t, i}^{s} = \mathds{1}_{[i \;= \; \mathrm{argmax}_j(p_{t, j}^{s})]}$.
During backward pass, it uses the continuous $p_{t, j}^{s}$, allowing backpropagation. At inference, largest index in $l_{t}^{s}$ is chosen.

\keypoint{Visual-Semantic Reasoning:}  The visual and semantic reasoning functions $\phi(\cdot)$ and $\omega(\cdot)$ are employed by Transformer module \cite{vaswani2017attention} that uses multi-headed self-attention mechanism to gather global context information. In brief, given key (K), query (Q) and value (V), attention is calculated as: $ \textit{Attention}( K, Q, V)  = \textit{softmax}(\frac{QK^\intercal}{\sqrt{dim}})V$. At each time step output $\phi_{t}(\cdot)$ and $\omega_{t}(\cdot)$, feature representation is enriched by information from remaining time-steps and thus long-range dependencies are modelled carefully.  Semantic reasoning module $\omega(\cdot)$ is pre-trained separately following BERT \cite{devlin2018bert} language model training topology. We mask out (also purposefully replace by erroneous instances) certain input time steps and force to predict masked token by a linear layer. This helps the model to learn better refining potential using text-only data in advance.

\keypoint{Dense and Residual Design:} 
Glimpse vector $g_{t}$ provide character localized visual information. For later stages $g_{t}^{s}$ is computed based on joint visual-semantic information so that more enriched representation can be extracted. To take advantage from multiple stages, we add a \emph{dense connection} \cite{huang2017densely} between computed current $g_{t}^{s}$ and preceding  $g_{t}^{<s} = \{g_{t}^{s-1}, \cdots, g_{t}^{0} \}$ as : $\mathrm{\mathbf{\overline{g_{t}^{s}} = W_{g}^{s} \otimes [g_{t}^{s}, g_{t}^{s-1}, \cdots, g_{t}^{0}]}}$, where,  $W_{g}^{s}$ is trainable parameter and implemented through $1\times1$ convolution.

To sum up, we get a differentiable semantic space represented by one-hot encoding as: $Y^{s-1} = \{y_{t}^{s-1}\}_{t = 1}^{T} $, where $y_{t}^{s-1} = \texttt{Gumbel-Softmax}(l_{t}^{s-1})$. Next, we calculate joint visual-semantic feature $\mu_{t}^{s}$ and successively $g_{t}^{s}$ is computed. Glimpse vector $g_{t}^{s}$ for $s\geq1$ is enhanced by dense connection to give $\overline{g_{t}^{s}}$. Now we update the hidden state of current stage decoder RNN by: $\mathrm{\mathbf{H_t^{s} =  {f_{rnn}^{s}}(H_{t-1}^{s}; \; [E(y_{t}^{s-1}), \;  \overline{g_{t}^{s}}, \; \mu_{t}^{s}])}}$. Excluding the final stage, we directly apply linear classifier to get: $\mathrm{P(y_{t}^{s}) = \texttt{softmax}(F_{cls}^s(H_{t}^{s}))}$. 

For the final stage, $s = S$, we add a residual connection \cite{he2016deep} between initial $H_{t}^{0}$ and final $H_{t}^{S}$using LayerNorm \cite{vaswani2017attention} as follows: $\overline{H_{t}^{S}} = \texttt{LayerNorm}(H_{t}^{S} + H_{t}^{0})$. The motivation aligns with original residual convolutional architecture \cite{he2016deep}, but here we integrate it to train deeper model with multiple attention decoder stages for text recognition. The final prediction is obtained as $\mathrm{\mathbf{P(y_{t}^{S}) = \texttt{softmax}(F_{cls}^S(\overline{H_{t}^{S}}))}}$. See Figure \ref{fig:Fig2} for more clarity. 

\setlength{\abovedisplayskip}{0cm}
\setlength{\belowdisplayskip}{0cm} 

\subsection{Learning Objective} \label{learning}
We accumulate cross-entropy loss from all stages of attentional decoder to train our text-recognition model. 
\begin{equation}\label{loss1}
L_{C} = -  \sum_{s = 0}^{S} \sum_{t=1}^{T} \hat{y}_{t}\cdot log P(y_{t}^{s}|H_{t}^{s})  
\end{equation}
where $\mathrm{\hat{Y} = \{ \hat{y_{1}}, \hat{y_{2}}, \cdots, \hat{y_{T}}\}}$ is the ground-truth label. Furthermore, we use additional \textit{auxiliary} linear classifier over character aligned individual visual and semantic features  $\vartheta_{t}^{s}$ and $\chi_{t}^{s}$ respectively, that are processed through reasoning modules. The next two losses could be thought of as an auxiliary loss driving towards better convergence that enrich individual character aligned feature with better visual and semantic information. This is given by: ${L_{V} = - \sum_{s = 1}^{S} \sum_{t=1}^{T} \hat{y}_{t}\cdot log P(y_{t}^{s}|\vartheta_{t}^{s})}$ and ${ L_{S} = - \sum_{s = 1}^{S} \sum_{t=1}^{T} \hat{y}_{t}\cdot log P(y_{t}^{s}|\chi_{t}^{s}) }$. The network is thus trained using : $\mathrm{L_{Total} = \lambda_{1}\cdot L_{C} + \lambda_{2}\cdot L_{V} + \lambda_{3}\cdot L_{S}}$, where $\lambda_{1}, \lambda_{2}, \lambda_{3}$ are weights decided empirically. 

\vspace{-.2cm} 
\section{Experiments}\label{sec:experiments}
\vspace{-.1cm}  
\keypoint{Datasets:} Following the similar approach described in \cite{ESIR2019,yang2019symmetry, baek2019wrong, cheng2018aon, shi2018aster, moran}, we train our model on synthetic datasets (without any further fine-tuning) such as SynthText \cite{SynthText} and Synth90k \cite{jaderberg2014synthetic}, which holds 6 and 8 million images respectively. The evaluation is performed without fine-tuning on datasets containing real images like: \noindent{\textbf{Street View Text (SVT)}}, \noindent{\textbf{ICDAR 2013 (IC13)}}, \noindent{\textbf{ICDAR 2015 (IC15)}}, \noindent{\textbf{CUTE80}}, \noindent{\textbf{SVT-Perspective (SVT-P)}}, \noindent{\textbf{IIIT5K-Words}}.
Street View Text dataset \cite{WangICCV2011} consists of 647 images, most of which are blurred, noisy or have low resolution. While ICDAR 2013~\cite{ICDAR2013} has 1015 words, ICDAR 2015 contains a total of 2077 images of which 200 images are irregular. CUTE80 \cite{CUTE80} offers 288 cropped high quality curved text images. SVT-Perspective \cite{SVT-P} presents 645 samples from side-view angle snapshots containing perspective distortion. IIIT5K-Words \cite{IIIT5K-Words} distinguishes itself by presenting randomly picked 3000 cropped word images.

\keypoint{Implementation Details:}  We use ResNet architecture from \cite{shi2018aster} with FPN heads having 256 channels in each multi-scale feature-maps. The kernel size of intermediate pooling layers is so adjusted that $B_{L}, B_{L-1}, B_{L-2}$ have spatial size of $4\times25$, $8\times25$, and $16\times50$ respectively. The hidden state size of two-layer encoder BLSTM and each decoder LSTM is kept at 256. Semantic $(\omega)$ and visual $(\phi)$ reasoning blocks consist of 2 stacked transformer units \cite{vaswani2017attention} with 4 heads and hidden state size 256. The hidden units in attention block is of size 128. A total of 37 classes are taken including alphatbets, numbers and end-tokens; with the maximum sequence length (N) set to 25. We use ADADELTA optimizer \cite{baek2019wrong} with learning rate 1.0 and batch size 32. We resize the image to 32x100 and train our model in a 11 GB NVIDIA RTX-2080-Ti GPU using PyTorch. We first warm-up using single stage attentional decoder for 50K iterations, and then train our proposed three-stage ($S=\{0,1,2\}$) attentional decoder (ablation on optimal stages in Sec. \ref{abla}) framework end-to-end, for 600K iterations with $\lambda_1, \lambda_2, \lambda_3$ set to 1, 0.1, 0.1 respectively. Please note that the first stage is fed with one-time step shifted ground truth label to accommodate teacher forcing in sequence modeling, however, later stages are fed with model's prediction from previous stage in order to learn the data driven refining strategy.  

\begin{table*}
\centering \label{main_table}
\caption{Comparison of proposed method with different state-of-the-art methods.}
\label{tab:comparison}
\vspace{-0.3cm}
\resizebox{2.1\columnwidth}{!}{
\begin{tabular}{c|c|c|c|c|c|c|c|l}
\toprule\toprule
\textbf{Methods} & \textbf{Year}  & \textbf{IIIT-5K} & \textbf{SVT} & \textbf{IC13} & \textbf{IC15} & \textbf{SVT-P} & \textbf{CUTE80} & \textbf{Remarks} \\
    \midrule
    Shi \etal \cite{Shi2015} & 2015 & 81.2 & 82.7 & 89.6 & - & 66.8 & 54.9 & \textbullet \;End-to-end trainable CNN + RNN + CTC. \\
    \midrule
    Lee \etal \cite{lee2016recursive} & 2016 & 78.4 & 80.7 & - & 90.8 & - & 42.7 & \textbullet \;Recursive CNN + RNN + Atten. decoder. \\
    Shi \etal \cite{shi2016rare} & 2016 & 81.9 & 81.9 & 88.6 & - & - & - & \textbullet \;Introduce rectification network for irregular images. \\
    \hline
    Cheng \etal \cite{cheng2017focusing} & 2017 & 87.4 & 85.9 & 93.3 & 70.6 & 71.5 & 63.9 & \textbullet \;Learning to focus on character centre, but needs char. location label.\\
    \midrule
    Liu \etal \cite{liu2018charnet} & 2018 & 83.6 & 84.4 & - & 60.0 & 73.5 & - & \textbullet \;Rectify the distortion at individual character level. \\
     Bai \etal \cite{bai2018edit} & 2018 & 88.3 & 87.5 & 94.4 & 73.9 & - & - & \textbullet \;Edit distance based seq. dissimilarity modeled to handle noisy characters. \\ 
    Liu \etal \cite{liu2018feature} & 2018 & 89.4 & 87.1 & 94.0 & - & 73.9 & 62.5 & \textbullet \;Leverage rendering parameters of synth. word image generation for training. \\
    Shi \etal \cite{shi2018aster} & 2018 & 93.4 & 93.6 & 91.8 & 76.1 & 78.5 & 79.5 & \textbullet \;Improved rectification network by Thin-Plate Spline.\\
    Cheng \etal \cite{cheng2018aon} & 2018 & 87.0 & 82.8 & - & 68.2 & 73.0 & 76.8 & \textbullet \;Four directional convolutional feature extraction for irregular images.\\
    \midrule
    Liao \etal \cite{liao2019scene} & 2019 & 91.9 & 86.4 & 91.5 & - & - & 79.9 &    \textbullet \; Segment individual character + discrete char. recog. and word formation. \\
    Yang \etal \cite{yang2019symmetry} & 2019 & 94.4 & 88.9 & 93.9 & 78.7 & 80.8 & 87.5 & \textbullet \;Models geometrical attributes of text for better images rectification. \\
    Li \etal \cite{li2019show} & 2019 & 95.0 & 91.2 & 94.0 & 78.8 & 86.4 & 89.6 & \textbullet \;Introduce 2D-attention to deal with irregular images. \\
    Baek \etal \cite{baek2019wrong} & 2019 & 87.9 & 87.5 & 92.3 & 71.8 & 79.2 & 74.0 & \textbullet \;Comparative study of different methods and insightful analysis. \\
    Zhan \etal \cite{ESIR2019} & 2019 & 93.3 & 90.2 & 91.3 & 76.9 & 79.6 & 83.3 & \textbullet \;Iterative image rectification. \\
    \midrule
    Litman \etal \cite{litman2020scatter} & 2020 & 93.7 & 92.7 & 93.9 & 82.2 & 86.9 & 87.5 & \textbullet \;Stacking more Bi-LSTM layers + gated fusion of visual-contextual feature. \\
    Qiao \etal \cite{qiao2020seed} & 2020 & 93.8 & 89.6 & 92.8 & 80.0 & 81.4 & 83.6 & \textbullet \;Tries to predict the word-embedding vector to initialise the state of decoder. \\
    Yu \etal \cite{yu2020towards} & 2020 & 94.8 & 91.5 & 95.5 & 82.7 & 85.1 & 87.8 & \textbullet \;Faster parallel decoding + semantic reasoning block (\emph{non-differentiable}).   \\
    \midrule
    \textbf{Our Baseline} (Stage-0) & - & 88.0 & 84.9 & 90.4  & 74.5  & 75.3  & 82.6 & \multirow{4}{*}{\shortstack[l]{\textbullet \;Joint visual-semantic reasoning through multi-stage decoding using \\ multi-scale feature maps and differential semantic space.}} \\
    \textbf{Our Baseline} (Stage-1) & - & 92.6 & 89.5 & 93.9 & 80.3 & 81.5 & 87.2 \\
    \textbf{Proposed} (Stage-2) & - & \textbf{95.2} & \textbf{92.2}  & \textbf{95.5} & \textbf{84.0}  & \textbf{85.7}  & \textbf{89.7} \\
    \textbf{Our Baseline} (Stage-3) & - & 95.2 & 92.1  & 95.5 & 83.6  & 85.5  & 89.6 \\
    \bottomrule
\end{tabular}
}
\vspace{-0.7cm}
\end{table*}

\subsection{Result Analysis and Discussion}\label{analysis}

Table \ref{tab:comparison} shows our proposed method to surpass SOTA methods by a reasonable margin. Every method's salient contributions are briefly mentioned there as well. 
In this section, we first describe the limitations of the existing or alternative (naive) designs and then illustrate (\emph{using} \textbf{IC15}) how and why all our design components/choices contribute towards superiority over others.

\begin{figure}[!hbt]
\begin{center}
  \includegraphics[width=\linewidth]{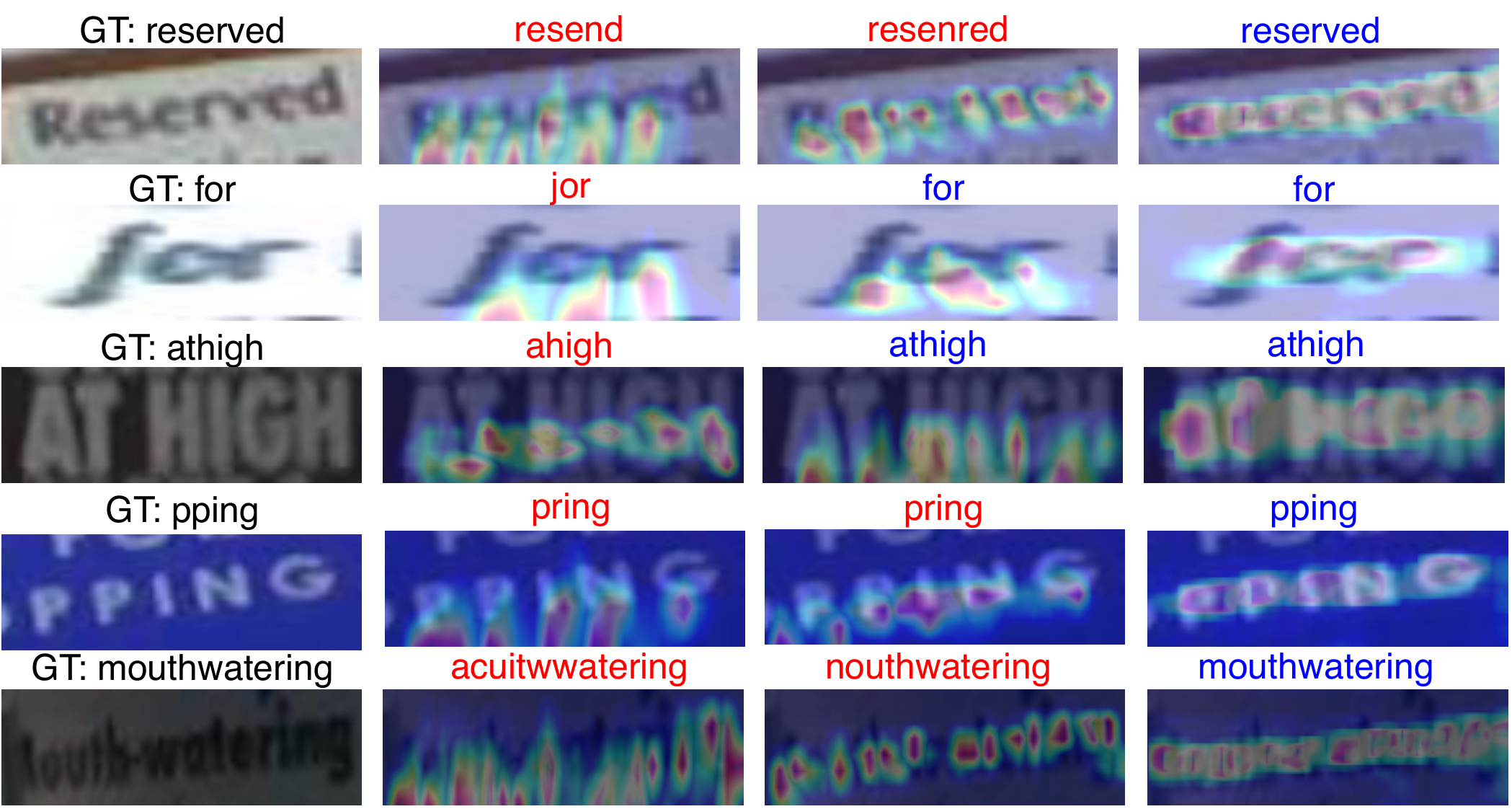}
\end{center}
\vspace{-.25in}
  \caption{Examples showing how joint visual-semantic information could help to recognise through refining over stages $(s=0 \rightarrow s=1 \rightarrow  s=2 )$, shown left to right.}
\vspace{-.25in}
\label{fig:fig3}
\end{figure}

\keypoint{\emph{[i] Limitation of previous attentional decoders}:} Existing methods relying on unidirectional auto-regressive attentional decoders exhibit a bottleneck, and its drawback becomes evident from the following scenario : An easily recognizable character present towards the end of a word would fail to provide any contextual semantic information towards recognizing some noisy character present earlier. We on the contrary let the first stage completely unroll itself. Thereafter the prediction of previous stage (even if certain time-step's character is incorrect) could be rectified in the subsequent stages using joint visual-semantic information. Although SCATTER \cite{litman2020scatter} stacks multiple BLSTM layers on the top of baseline design from ASTER \cite{shi2018aster}, both methods lack semantic reasoning as they barely enrich \emph{visual} feature encoding. Examples from our stage-wise decoder are shown in Figure \ref{fig:fig3}.
 
\noindent \textbf{\emph{[ii] Significance of Differentiable Semantic Space:}} Improving semantic reasoning for better text recognition was only considered by \cite{yu2020towards} and \cite{qiao2020seed} among all SOTA methods. Although Qiao \etal \cite{qiao2020seed} proposed to use word embedding, such technique relies on semantic meaning of a word instead of the required character sequence. For example, the word ``table" and ``chair", although semantically related have character combinations that are way-apart. Therefore, we emphasise on modelling character sequences instead, to help recognize a noisy character based on two-way information passing. Even though Yu \etal \cite{yu2020towards} took this direction to some extent, their non-differentiable semantic-reasoning block imposes a significant limitation. We alleviate that with the help of gumbel-softmax \cite{jang2016categorical} to develop a differentiable semantic space and allow learning of multi-stage semantic reasoning. While the use of \emph{teacher forcing} for later stages by feeding ground-truth label for training multi-stage decoder might seem an alternative, empirical evidence suggests otherwise. The third stage decoder obtains {$74.4\%$} accuracy as compared to $74.5\%$ accuracy (on IC15) in first stage -- no practical gains. Another straight-forward way is to use \emph{straight-through estimator} \cite{bengio2013estimating}, which simply copies gradients from argmax output to the next input. However, this results in significant instability where later stage performance drops by $3.9\%$ to {$80.1\%$} due to discrepancies between forward and backward passes resulting in much higher variance than gumbel-softmax \cite{jang2016categorical}. 

\noindent \textbf{\emph{[iii] Why not directly use logits instead of gumbel-softmax for semantic reasoning}:} Feeding logits (probability distribution over character vocabulary prior to $argmax$) from a previous stage to the next, is a reasonable argument that would make everything differentiable and eliminate the need for gumbel-softmax. However, it is important to remember that characters are \emph{discrete} tokens \cite{baek2019wrong}. Using logits requires one to replace character embedding layer $\mathrm{E(\cdot)}$ by a simple FC layer. Unlike  $\mathrm{E(\cdot)}$ that picks up a particular row of a trainable matrix based on discrete one-hot vector, a FC layer will give varying representations for the same character sequence based on the \emph{confidence} of predictions. We confirm this hypothesis of sub-optimality empirically, as results drop from $84.0\%$ to $82.8\%$.

\keypoint{\emph{[iv] Why use top-down attentional decoder}:} 
While low resolution and semantically strong features are good for classification, tasks requiring focus in local regions, such as object detection and semantic segmentation, benefit even further when combined with high-resolution semantically weak features found in shallower regions of a feature extractor \cite{cai2016unified}. Although our first stage is similar to a basic attentional decoder focusing on feature map of the last layer to benefit from rich semantic information, that is more invariant to distortion, later stages (refining stages) combine higher resolution feature-map from preceding layers. This not only handles varying character size, but also verifies prior prediction by exploiting joint information between high resolution feature and previous predictions to guide the \emph{refining} process. This hypothesis is verified by contradiction, using high-resolution semantically weak feature $B_{L-2}$ in $s=0$ and lower resolution semantically strong features in later stages $s>1$. We observe performance collapses to {$72.1\%$} in IC15 dataset due to inability of high resolution semantically weak features to output the initial estimates.

\noindent \textbf{\emph{[v] Significance of self-attention based Joint Visual-Semantic Reasoning}:}
To emulate \emph{human-like} inference, self-attention based \cut{joint visual-semantic} reasoning functions allow two way information passing across visual and semantic spaces to obtain a joint visual-semantic context. Its significance \cut{of such a design choice} could be empirically understood by removing the visual reasoning block and modifying the architecture accordingly, which drops result by $2.9\%$. A similar drop of $4.8\%$ was observed when the semantic reasoning block was removed. \cut{, leading to a $4.8\%$ performance drop} On removing both we observe $77.1\%$ accuracy -- a significant drop of $6.9\%$ from our method (Table \ref{tab:constraints}). 

\noindent \textbf{\emph{[vi] Do multi-scale (resolution) feature maps help?}} We empirically validate this by excluding multi-scale feature maps and use $B_L$, instead of $B_{L-s}$, to calculate $g^{s}_t$ at every stage $s$. Such modification drops performance by $2.7\%$ (against ours), to {$81.3\%$}, which highlights the contribution of multi-scale feature maps in our method. 

\noindent \textbf{\emph{[vii] Comparison with alternative  \emph{multi-scale} attentional decoder designs}:} 
In text recognition, the only other work realising importance of multi-scale information is by Wan \etal \cite{wan2020vocabulary}, where pyramid pooling was used.\cut{ to achieve this goal. }
Here visual feature maps from different spatial resolutions were concatenated, which eventually harmed downstream tasks owing to the large semantic gaps between such feature maps.
Consequently, we introduce lateral connections following Feature Pyramid Networks \cite{lin2017feature}, semantically strengthening high-resolution levels for superior performance.
Simply employing pyramid pooling for all stages $s=\{0, 1, 2\}$ however, drops performance by $2.1\%$ (against ours) to $81.9\%$ .

\keypoint{\emph{[viii] Significance of Dense and Residual Connections:}}
Beside improving visual information flow in the forward pass, the residual connection between initial $H_t^{0}$ and final $H_t^{S}$ ensures efficient gradient flow in visual feature networks, accelerating convergence of the whole network. Furthermore, the dense connection is used to adaptively learn a more discriminative glimpse vector by combining its features from preceding stages with the current one, thus stabilising the training of multi-stage multi-scale attentional decoder. Removing dense connection ($g_t$ calculation) decreases the performance by {$1.6\%$}, and removing residual connection decreases it by {$1.3\%$}. On removing both we get an even larger drop of {$1.9\%$}. Faster training is observed while using both dense and residual connections.
 
\keypoint{\emph{[ix] Significance of Multiple Constraints:}} 
We design experimental setups (Table \ref{tab:constraints}) that reveal the following observations: (a) imposing loss $L_{C}$ only in the last stage harms the model, resulting in  {$73.1\%$} accuracy. We attribute this to the poor gradient flow across stages. (b) Adding multi-stage $L_{C}$ loss results in  $77.1\%$ accuracy, performing closer to the proposed method. (c) Adding visual-semantic constraints $L_{V}$ and $L_{S}$ finally gives the best performance of  $84.0\%$. This shows multi-stage constraint is vital for training and convergence. The intuition behind multiple constraints sources from multi-task learning, which ensures better convergence, thus enriching individual character aligned feature, with better visual-semantic information.

\keypoint{\emph{[ix] Varying training data size:}} Following \cite{luo2020learn}, we also vary the training size and evaluate our proposed framework compared to single stage baseline and Yu \etal \cite{yu2020towards} in Table \ref{tab:constraints}. Significant overhead at low data regime brings the superiority to our proposed method over others. 





\setlength{\tabcolsep}{1.8pt}
\begin{table}[]
    \centering
    \caption{ \doublecheck{(Left) Effect of multiple constraints on IC15. (Right) Varying training data size. 
     $L_{C}'$: Last stage only,  $L_{C}$: Multi-Stage, GAP: WRA margin against final performance.}}
    \footnotesize
    \vspace{-0.3cm}
    \begin{tabular}{ll}
        \begin{tabular}{cccccc}
            \hline
            $L_{C}'$ & $L_{C}$ & $L_{V}$ & $L_{S}$ & IC15 & GAP\\
            \hline
            \checkmark & - & - & - & 73.1 & 10.9 \\
            - & \checkmark & - & - & 77.1 & 6.9 \\
            - & \checkmark & \checkmark & - & 79.2 & 4.8 \\
            - & \checkmark & - & \checkmark & 81.1 & 2.9 \\
            \midrule
            - & \checkmark & \checkmark & \checkmark & 84.0 & - \\
           \hline
        \end{tabular}
        &
        \begin{tabular}{ccccc}
            \hline
            \multirow{2}{*}{Method} & Syn & Syn & Syn & Syn\\
            & 10K & 50K & 100K & 1M \\
            \hline
           
            Yu \etal \cite{luo2020learn} & 21.7 & 37.7 & 51.2 & 67.4 \\
            Luo \etal \cite{yu2020towards} & 13.3 & 32.1 & 47.3 & 63.7 \\
             Baseline (s=0) & 9.9 & 27.2 & 44.9 & 62.3 \\
            \midrule
            Proposed & 25.3 & 41.5 & 56.4 & 73.1 \\
           \hline
        \end{tabular}
    \end{tabular}
    \vspace{-0.4cm}
    \label{tab:constraints}
\end{table}

\vspace{-0.05cm}
\subsection{Further Analysis and Insights}\label{abla}
\vspace{-0.05cm}

\keypoint{\emph{[i] Design of Visual-Semantic Reasoning Module:}} One can capture two-way visual semantic information using (a) Bi-LSTM (b)  Transformer \cite{vaswani2017attention} with multi-headed self-attention mechanism. \cut{Efficacy of each variant is understood empirically } Table
\ref{tab:visual-semantic-LM} shows Transformer to outperform LSTM by $1.3\%$. Furthermore, pre-training global semantic reasoning module $\omega(\cdot)$ using BERT \cite{devlin2018bert} like training topology, scores $0.9\%$ higher accuracy than without it.

\setlength{\tabcolsep}{6pt}
\begin{table}[]
    \centering
    \caption{Significance of joint visual-semantic reasoning module and comparison with Language Models (LM).}
    \footnotesize
         \vspace{-0.3cm}
    \begin{tabular}{ccc}
        \hline
        \textbf{Methods} & IC15 & GAP \\
        \hline
        Our Baseline (Stage-0) + LM-shallow & 74.3 & 9.7 \\
        Our Baseline (Stage-0) + LM-deep & 75.9 & 8.1 \\
        Joint Visual-Semantic using LSTM & 81.8 & 2.2 \\
        Joint Visual-Semantic using Transformer & 83.1 & 0.9 \\
        \hline
        Transformer with Pre-Training Semantic reasoning & 84.0 & - \\
        \hline
    \end{tabular}
    \vspace{-0.2cm}
    \label{tab:visual-semantic-LM}
\end{table}
 
\keypoint{\emph{[ii] Weight sharing across stages:}} The stage-wise attentional decoder has five trainable modules, $\mathrm{F_{cls}(\cdot)}$, $\mathrm{E(\cdot)}$, $\mathrm{\Phi}(\cdot)$, $\mathrm{\omega(\cdot)}$ and $\mathrm{\psi(\cdot)}$, whose weights can either be shared across stages or have a separate model for each stage. Using separate weights achieves $82.5\%$ accuracy on IC15, whereas sharing across stages results in $82.3\%$. Interestingly using a separate $\mathrm{F_{cls}, \psi}$, and shared $\mathrm{E(\cdot),\; \Phi, \;\omega}$ gives $84.0\%$, a $1.7\%$ rise, in contrast to sharing all weights -- probably because sharing parameters which are not \emph{stage dependent} reduce model complexity and has better optimization.

\noindent \textbf{[iii] \emph{Computational Analysis:}} Each stage needs to unroll itself completely, before the next starts processing. Hence, the performance gain comes at a cost of extra computational expenses (analysis in Table \ref{tab:computational}), which is reasonable given the superior performance over strong baselines. \cut{A thorough ablative study is done on complexity analysis in Table \ref{tab:computational}.} Even so, we experimented with ResNet-101 as a backbone feature extractor, having similar number of parameters and flops to ours. This naive stacking of multiple-layers lags by 8.9\%, which accredits our gain to our novel design choice.

\begin{table}[]
    \centering
    \caption{Computational analysis of the proposed method.}
    \footnotesize
     \vspace{-0.3cm}
    \begin{tabular}{ccccc}
        \hline
        \textbf{Method} & GFlops & Params & CPU & IC15 \\
        \hline
        Our Baseline (Stage-0) & 15.3 & 38M & 16.38ms & 74.5 \\
        Proposed Method & 22.5 & 44M & 26.31ms & 84.0 \\
        \hline
    \end{tabular}
    \vspace{-0.6cm}
    \label{tab:computational}
\end{table}

\noindent \textbf{\emph{[iv] Comparison with SOTA Language Model:}}  We compare our framework with state-of-the-art Language Modeling (LM) based post-processing techniques based on librispeech text-corpus. Based on \cite{gulcehre2015using} we adopt two techniques: (a) Shallow Fusion that results in $74.3\%$ and (b) Deep Fusion giving $75.9\%$ accuracy on IC15 (Table \ref{tab:visual-semantic-LM}).

\noindent \textbf{\emph{[v] Optimum Stages:}} The optimal value for the number of stages $s$ is found empirically on IC15. For $s=1$ we have $80.3\%$ accuracy that improves at $s=2$ to give $84.0\%$, but saturates at $s=3$ giving $83.6\%$. Hence we consider $s=2$ to be optimal. This performance saturation could be attributed to vanishing gradient problem which is addressed via residual/dense connection, but still persists to some extent. Also, for $s > 2$, the joint visual-semantic information might reach its optimum, where the result saturates. Please refer to \emph{supplementary material} as well.  

\vspace{-0.1cm} 
\section{Conclusion}
\vspace{-0.1cm}


We propose a novel joint visual-semantic reasoning based multi-stage multi-scale attentional decoding paradigm. The first stage predicts from visual features, followed by refinement using joint visual-semantic information. We further exploit Gumbel-softmax operation to make visual-to-semantic embedding layer differentiable. This enables backpropagation across stages to learn the refining strategy using joint visual-semantic information. Experimental results indicate the superior efficiency of our model.

{\small
\bibliographystyle{ieee_fullname}
\bibliography{egbib}
}

\end{document}


\title{\vspace{-8mm} \large {Supplementary for \\ Joint Visual Semantic Reasoning: Multi-Stage Decoder for Text Recognition}}  

\maketitle

\section*{Additional Explanations}

\noindent \textbf{$\bullet$ Optimal number of stages:} In any multi-stage network, it is quite natural for performance to saturate after a few stages. For us, it saturates after s = 2. It is well accepted in parallel literature that adding more layers to convolutional network does not necessarily improve the model's performance even if representation power is increased theoretically. Moving forward, ResNet was designed to handle very deep networks, however even for ResNet \cite{karayev2012timely}, adding more layers often saturates or even lowers performance in various tasks. Therefore, our empirical observation is not a surprise and is aligned with the findings of other parallel works. This performance saturation could be attributed to vanishing gradient problem which is addressed via residual/dense connection, but still persists to some extent. Also, for s $>$ 2, the joint visual-semantic information might reach its optimum, where the result saturates.

\vspace{0.1cm}




\noindent \textbf{$\bullet$ Use of Gumbel-Softmax and its novelty:} Designing of Gumbel-Softmax \cite{jang2016categorical} is not actually our novelty. We deal with the idea that in order to exploit the semantic information for text recognition, prediction should be refined in a stage-wise manner. However, characters are discrete tokens, thus we emulate Gumbel-Softmax operation to enable gradient flow across-stages. The major contribution here is how we design a multi-stage framework that is unrolled in a stage-wise manner and meets the objective of visual-semantic reasoning tailor made for text recognition. 



\noindent \textbf{$\bullet$ Latency of the recognition modules:} Any multi-stage framework would have at least had a bit higher latency. \cut{However, had your argument been legitimate, there shouldn't have been any multi-stage frameworks in the computer vision literature --  this is not the case.} For instance, convolutional pose machine \cite{gao2019startnet} being a multi-stage framework is widely accepted by the community for pose estimation. 
Nevertheless, we perform a study where we use a ResNet-101 as backbone feature extractor to make the number of parameters and flops of baseline model very close to ours. However, the performance still lags by $8-9\%$ compared to ours. Therefore, our performance gain is not tied to higher computational cost as any naive stacking of multiple layers can not attain any performance gain. We conclude that well-motivated and careful designs (\emph{stage-wise unrolling}) lead to this performance boost. 


\noindent \textbf{$\bullet$ Justification of the multiple constraints:} The core motivation of multiple constraints came from multi-task learning, where multiple tasks are constrained to accelerate the learning process and reduce the training time. Similarly, the two losses $L_V$ and $L_S$ are used as auxiliary losses for driving towards better convergence that enriches individual character aligned feature, with better visual and semantic information. 






\noindent \textbf{$\bullet$ Why the proposed method has reasoning capability?}
\\Let's take a word image with annotation `aeroplane'. In case of single stage attentional decoder, if the model wrongly predicts `n' instead of `r',  `ae\textcolor{red}{n}' would have adverse effect on the rest of the prediction, and due to single stage prediction it would have no choice to refine the prediction. Furthermore, while predicting the first few characters, it has almost negligible semantic context information.
If we unroll the prediction in a stage wise manner, and a character is predicted wrongly, like `ae\textcolor{red}{n}oplane', it captures the semantic information from previous prediction. This aids in refinement during the later stages coupled with visual information (Eq. 3). Speaking intuitively, previous stage prediction `ae\textcolor{red}{n}oplane' could help refine `n' to `r' exploiting semantic reasoning. 


\noindent \textbf{$\bullet$ Background on Transformer \cite{vaswani2017attention} used for Visual-Semantic Reasoning:} A transformer network consists of encoder blocks, each comprising several layers of multi-head attention \cite{vaswani2017attention} followed by a feed forward network.
The encoder is usually designed as a stack of N identical layers, each of which has two sub-layers, a multi-head self-attention (MHA) and a feed-forward network (FFN). Contrary to traditional sequence modeling methods (RNN/LSTM) which learns the temporal order of current time steps from previous steps (or future steps in bidirectional encoding), the attention mechanism in transformers allows the network to decide which time steps to focus on to improve the task at hand. A single-head self-attention (SHA) module $\texttt{attn}(Q,K,V)$ is defined as:
\begin{equation}
    \texttt{attn}(Q,K,V) = \texttt{softmax}(\frac{QK^T}{\sqrt{d_k}})V
\end{equation}
Accordingly other network components are defined as:
\begin{equation}
\label{mha}
\begin{aligned}
    \mathrm{SHA}_i(Q,K,V) &= \texttt{attn}(QW^Q_i,KW^K_i,VW^V_i)  \\
    \mathrm{MHA}(Q,K,V) &= \texttt{concat}(\mathrm{SHA}_1, ...,\mathrm{SHA}_h)W^O 
\end{aligned}
\end{equation}
\begin{equation}
    \mathrm{FFN}(x) = \max(0, xW_1 + b_1)W_2 + b_2
\end{equation}
{where Q, K and V are \emph{Query}, \emph{Key} and \emph{Value} respectively, $\mathrm{SHA}_i$ is the $i$-th attention head. FFN ($\cdot$) is composed of two linear transformations with ReLU in between where {$W_{(\cdot)}^{(\cdot)}$ are learnable matrices.} Around each of the sub-layers, a residual connection with layer normalization is placed. Therefore, the output of each sub-layer is $\texttt{LayerNorm}(x+\texttt{sublayer}(x))$.}
The decoder is also designed as a stack of N identical layers following \cite{vaswani2017attention} with sub-layers, -- \textit{two} MHA sub-layers followed by 1 FFN. The additional MHA sub-layer performs multi-head \textit{self-attention} on the encoder's output. Similar to encoder stack, residual connections and layer normalizations are applied to each sub-layer as well. 

{\small
\bibliographystyle{ieee_fullname}
\bibliography{egbib}
}

